\documentclass[conference]{IEEEtran}
\IEEEoverridecommandlockouts

\usepackage{cite}
\usepackage{graphicx}

\usepackage{float}
\usepackage{hyperref}
\usepackage{booktabs}

\usepackage{threeparttable}
\usepackage{amsmath, amsfonts, amsthm, amssymb, mathtools, xparse, stackengine}
\usepackage{footnote}
\makesavenoteenv{threeparttable}
\usepackage{tablefootnote}
\usepackage{algorithmic, algorithm}
\stackMath
\usepackage{graphicx,float}
\usepackage{subfigure}
\usepackage{multirow}
\usepackage{makecell}

\begin{document}

\title{A pruning method based on the dissimilarity of angle among channels and filters}

\author{
    \IEEEauthorblockN{1\textsuperscript{st}Jiayi Yao\IEEEauthorrefmark{1}, 2\textsuperscript{nd}Ping Li\IEEEauthorrefmark{1}\IEEEauthorrefmark{2}, 3\textsuperscript{rd}Xiatao Kang\IEEEauthorrefmark{1}, 4\textsuperscript{th}Yuzhe Wang\IEEEauthorrefmark{1}}\\
    \IEEEauthorblockA{\IEEEauthorrefmark{1}Changsha University of Science and Technology, China}
    \IEEEauthorblockA{\IEEEauthorrefmark{2}Hunan Provincial Key Laboratory of Intelligent Processing of Big Data on Transp, China}
    \IEEEauthorblockA{\tt email: yjy\_toka@163.com, lping9188@163.com, kangxiatao@gmail.com, zheeln@qq.com}
}
\maketitle

\begin{abstract}
Convolutional Neural Network (CNN) is more and more widely used in various fileds, and its computation and memory-demand are also increasing significantly. In order to make it applicable to limited conditions such as embedded application, network compression comes out. Among them, researchers pay more attention to network pruning. In this paper, we encode the convolution network to obtain the similarity of different encoding nodes, and evaluate the connectivity-power among convolutional kernels on the basis of the similarity. Then impose different level of penalty according to different connectivity-power. Meanwhile, we propose Channel Pruning base on the Dissimilarity of Angle (DACP). Firstly, we train a sparse model by GL penalty, and impose an angle dissimilarity (AD) constraint on the channels and filters of convolutional network to obtain a more sparse structure. Eventually, the effectiveness of our method is demonstrated in the section of experiment. On CIFAR-10, we reduce 66.86\% FLOPs on VGG-16 with 93.31\% accuracy after pruning, where FLOPs represents the number of floating-point operations per second of the model. Moreover, on ResNet-32, we reduce FLOPs by 58.46\%, which makes the accuracy after pruning reach 91.76\%. Our code is made public at: \url{https://github.com/kangxiatao/prune_tf2_master}.

\end{abstract}


\section{Introduction}

Convolutional Neural Networks (CNNs) bring excellent performance, which makes great achievements in image processing, speech recognition etc. Nevertheless, for devices with limited computation and memory, such as mobile embedded devices, even with the latest high-efficiency architecture , the size and over-parameterization of its model are still burdensome to deploy  on neural networks , which will also affect the combination of CNNs and many traditional industries. Thus, network pruning provides the possibility and necessity for neural network compression.

Early studies have proposed many methods of network compression. Such as weight quantization \cite{bib1}, low-rank decomposition \cite{bib2}, knowledge distillation \cite{bib3}, pruning \cite{bib4,bib5,bib6,bib7}. In this paper, we focus on network pruning. After simplification, the network not only reduces amounts of computation, but also improves its generalization capability. Early pruning methods are weight pruning mostly \cite{bib8}, which is practiced by directly deleting unimportant parameters. However it will cause unstructured sparsity of the model. Hence, by removing filters \cite{bib4,bib5,bib6,bib9,bib10} or channels \cite{bib11} from the convolutional kernels, it will become a more widespread choice to leave a compact and coherent model by filter-level pruning.

At present, a great deal of pruning techniques base on redundancy and importance. Redundancy-based pruning usually counts the number of filters/channels as its redundancy. Important-based pruning depends on different definitions of importance for pruning. Some of them attach importance to norms, and nevertheless, this "norm-only" pruning has great limitations. For instance: (1) Remove the weight of the filters according to the norm value \cite{bib12}. (2) Directly remove part of the filters \cite{bib13}. (3) Sparse networks are left by pruning connections with redundancy or low weights \cite{bib7}. Some of them do not take into account the diversity of distribution of filters among layers, so they will have a negative impact on accuracy \cite{bib11}. Meanwhile, there are many pruning methods based on ``smaller-norm-less-informative''. In fact, small values are not equal to unimportant values \cite{bib13}. Filters with small norm values in the front may play an important role behind \cite{bib11}.

\noindent
\begin{figure}[htbp]
    \subfigure[L2-norm penalty]
    {
        \begin{minipage}[t]{.46\linewidth}
            \centering
            \includegraphics[scale=0.5]{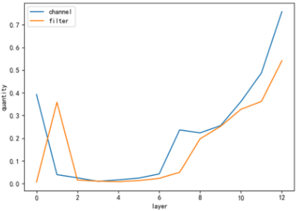}
            \label{1a}
        \end{minipage}
    }
    \subfigure[GL pruning]
    {
         \begin{minipage}[t]{.46\linewidth}
            \centering
            \includegraphics[scale=0.5]{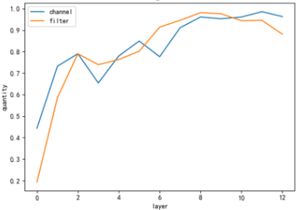}
            \label{1b}
        \end{minipage}
    }
\caption{The connectivity-power of channels and filters on VGG-16. Horizontal axis represents the number of layers, and vertical axis represents the similarity between channels (filters) of each layer. Blue lines denote channels and yellow lines denote filters. Inspired by the performance ability of related models, we try to increase the difference and improve their performance ability.}
\label{figure1}
\end{figure}

We try to prune the model while maintaining its performance, and encode the convolutional layer into a hypercube network. When the m-base strings of the encoding nodes differ by at most one bit, they are defined as a pair of adjacent nodes. As shown in Fig. \ref{figure1}, we calculate the mean of adjacent nodes to evaluate the connectivity-power of channels and filters between 2D convolutional kernels. In addition, we attempt to impose different LASSO penalties on the model to make the model structured sparsity. However, LASSO can only zero out the parameters of a single feature. Features appear in the form of groups, and a whole group of parameters need to be zeroed out at the same time. To solve this problem, Yuan \cite{bib14} proposed Group LASSO (GL) in 2006. As can be seen in Fig. \ref{figure1}\subref{1b}, after compression the connectivity-power of filters and channels in the middle layer increases significantly. Meanwhile, the performance of model is weak. Among them, the connectivity-power is the similarity between the channels/filters of each layer. The dispersion of channels in each layer is measured to facilitate dynamic pruning, and the filters are pruned in the same way \cite{bib13}.

\begin{figure}[htbp]
    \subfigure[GL constraints]{
        \begin{minipage}[t]{.44\linewidth}
            \includegraphics[scale=0.074]{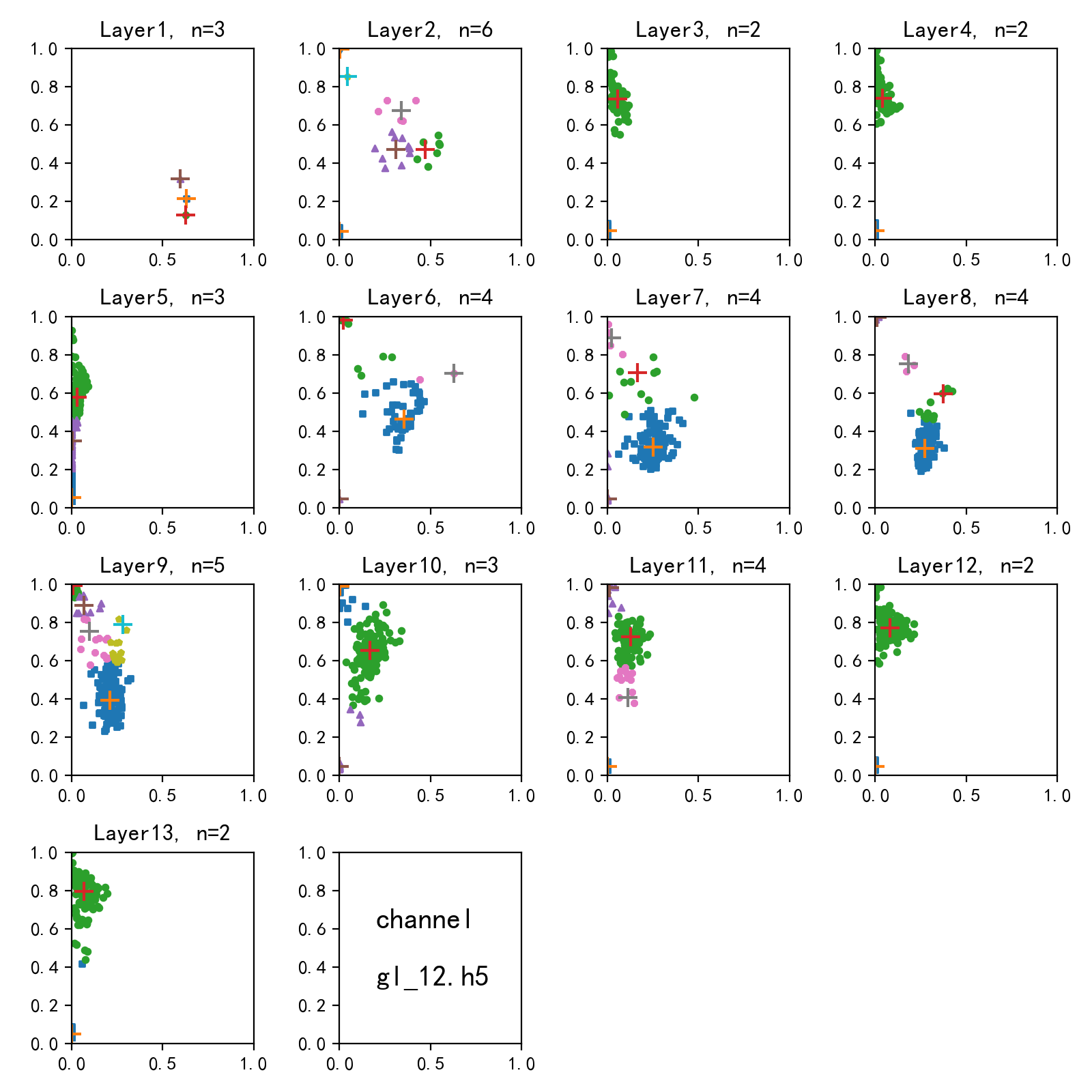}
            \label{2a}
        \end{minipage}
    }
    \subfigure[AS constraints]{
         \begin{minipage}[t]{.44\linewidth}
            \includegraphics[scale=0.074]{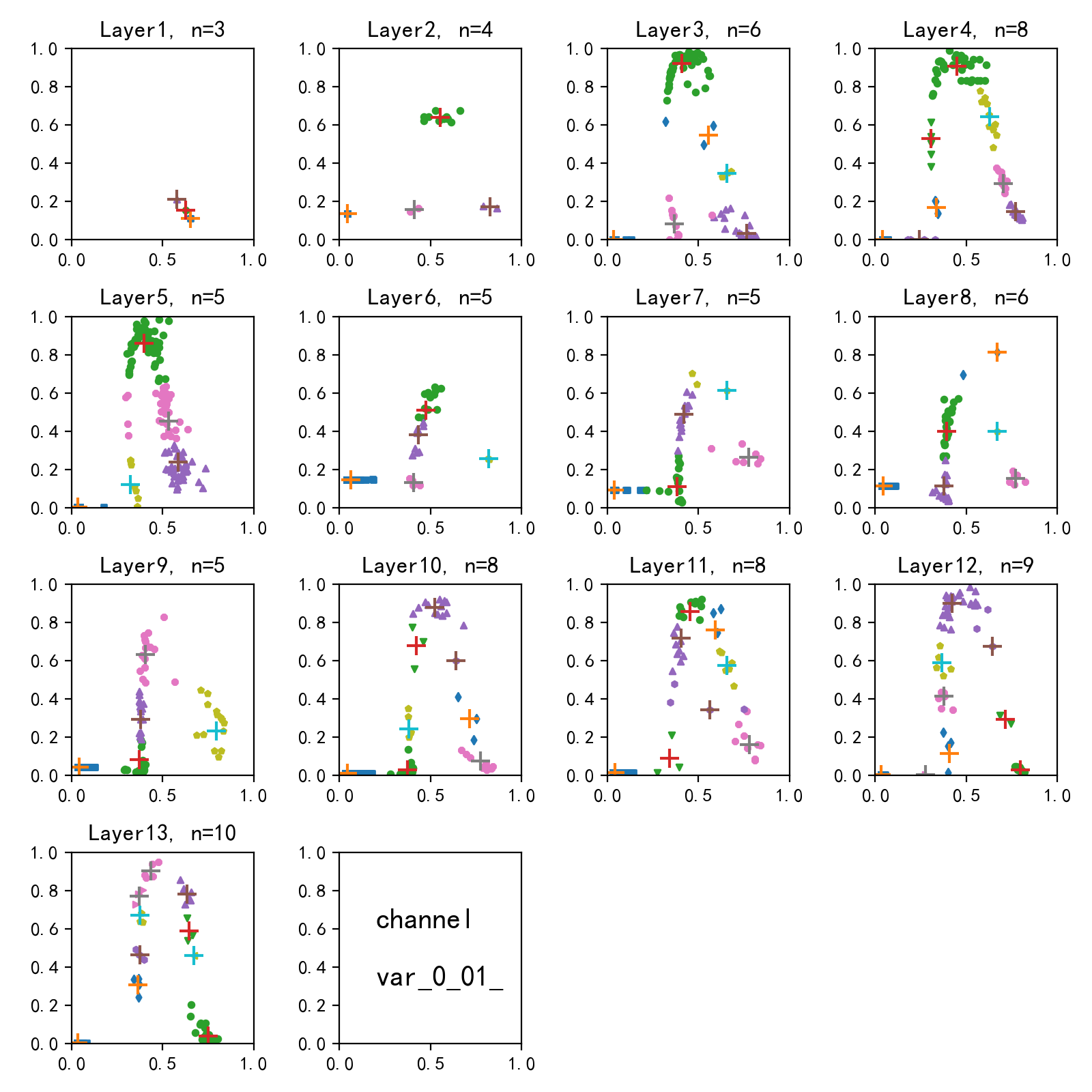}
            \label{2b}
        \end{minipage}
    }
\caption{Clustering analysis of GL constraint and AS constraint. A comparative experiment is conducted on VGG-16, with a total of 13-layers CNN, and each coordinate axis represents a single layer. The horizontal axis indicates the Euclidean distance between the channels or filters to the origin, and the vertical axis indicates AS between the channels/filters and their mean vector. The Euclidean distance and AS value are standardized between $[0,1]$. Different colors represent different categories. n is the number of categories, and `+' represents the center point of category.}
\label{figure2}
\end{figure}

In order to solve the problems above, we propose a novel model penalty method, Channel Pruning Based on the Dissimilarity of Angle (DACP). Firstly, a sparse model is trained by GL penalty, and then the Similarity of Angle (AS) constraint is constructed on the channels/filters of the CNN to increase the difference between the channels/filters and further screen out important filters to obtain a more sparse structure. As can be seen in Fig. \ref{figure2}, DACP separates the similarity between channels/filters and indirectly changes its norm value combined with GL constraint. So that filters with small but important values can be retained and the performance of convolutional kernels can be enriched. Besides, channels/filters with larger but similar values are penalized. In this case, it not only improves the generalization but also makes the model sparse.

As for filter selection, DACP is somewhat similar to filter pruning \cite{bib4}, but the latter is obtained by penalty-induced-sparsity and considering that filters close to geometric median have more redundancy. As for processing, it is resemblance to Dynamic Pruning \cite{bib15}, which increases the connectivity-power of the model and further inhibits the connectivity-power of the redundancy. Besides, since our method is based on GL to penalize the sparsity parts, our method does not need to achieve with extra pre-training and to achieve a compact model structure in the end of training. Above all, traditional pruning methods can also be applied to our trained models.

In this paper, we make a new assumption and propose ours for the existing limitations and shortcomings, which solves the disadvantages of poor performance and  generalization. Finally, for verifying the effectiveness of DACP, we use multiple image recognition datasets on multiple network architectures. We compare it with a variety of widely used pruning methods, and describe in detail in the third section of the paper.

\section{Related Work}

As mentioned above, previous works can be divided into unstructured pruning and structured pruning. Unstructured pruning \cite{bib7,bib8,bib16,bib17} mainly includes weight pruning and neuron pruning, which results in unstructured sparse model. Structured pruning \cite{bib4,bib5,bib6,bib9,bib18,bib10} mainly contains filter pruning and channel pruning. The compact network obtained after pruning can maintain the original structure, but usually leads to a significant decrease in accuracy \cite{bib10}.

\subsection{Unstructured Pruning}

Originally, LeCun \cite{bib8} trimmed all unimportant weights in the network to improve the accuracy and generalization of the network by considering the weight parameters in the network as a single parameter. Later, Hassibi \cite{bib16} proposed OBS technique based on LeCun's \cite{bib8} to improve the restoration of the weight updating. Compared with the former, OBS can prune more weight as the same error. Recently, pruning focuses on reserving important connections in network according to norm value \cite{bib7}, so as to reduce the number of parameters and computation consuming of the model without affecting the final accuracy of the network \cite{bib7}. Neuron pruning sets the row/column of the matrix to zero, and the size of the matrix does not change. Hu \cite{bib17} defined ``average percentage of zero'' to measure the number of zero activated in each filter, and regards its neurons as redundant and prunes as a whole.

\subsection{Structured Pruning}

Contrary to unstructured pruning, sparsity brought by structured pruning is regular, and will obtain structured model eventually. So it is more feasible currently. Filter pruning and channel pruning belong to structured pruning. Previous pruning based on norm value include the filters pruning on account of L1-norm \cite{bib10}, L2-norm \cite{bib9} and Lp-norm \cite{bib5}. Besides, there are also pruning methods based on redundancy, such as He's \cite{bib4}. However, these methods always only lay emphasis on how to prune in the same layer, without noticing the relationship among different layers. Aiming at the limitations above, Luo \cite{bib6} attaches more importance to the relationship among layers, and he supposed that statistics calculated in next layer are used as the benchmark for pruning. Further, Xie \cite{bib18} proposed the concept of Extended Filter Group (EFG), which solves the training of filters of the current layer and corresponding channels of the next layer according to the penalty of EFG, and conducted induced-sparsity of the model. Our approach sorts out the above problems and put forward a more effecitve method to achieve better performance.

\section{Method}

Our pruning method can be summarized as the following steps: (1) Impose GL penalty on the model until what we set. (2)Reduce the GL penalty, and add the penalty of AD in channels until the final convergence. (3) Prune the channels according to the norm value of 3D filter.In this section, we will elaborate the setting of penalty terms and the learning process of structured sparsity.

\subsection{Preliminaries}

We formally introduce the symbols and notations in this subsection. We assume that a neural network has $L $ layers, and in the i-th layer we parameterize CNN as $W=\left\{W^{1}, W^{2}, \ldots, W^{L}\right\}$ as $\left\{W^{(\mathrm{i})} \in \mathbb{R}^{\mathit{k}  \times \mathit{k}  \times \mathit{c}^{\mathit{l}} \times \mathit{n}^{\mathit{l}}}\right\}$, where $\mathit{k}$ denotes kernel size, $\mathit{c}^{\mathit{l}} $ and $\mathit{n}^{\mathit{l}} $ denote $\mathit{l}$-th layer's number of channels and filters respectively. In addition, the convolutional kernel of the $\mathit{n}$-th filter in the $\mathit{c}$-th channel at the $\mathit{l}$-th layer is expressed as $W_{c,n}^{l}$.

\subsection{Convolutional Kernels Grouping}

First we divide the convolutional kernels of each layer into two groups: channel-filter and 3D-filter, and then impose L2-norm penalty on each group in the objective function:

\begin{equation}
    R_{g}=\sum_{l=1}^{\mathcal{L}}\left(\sum_{c=1}^{c^{l}}\left\|W_{c, n^{l}}^{l}\right\|_{2}+\sum_{n=1}^{n^{l}}\left\|W_{c^{l}, n}^{l}\right\|_{2}\right)
    \label{equ1}
\end{equation}

Indeed we can also design more complex groups to obtain better sparse effect in the training process, but the penalty based on AS only needs group constraints to work simply.

\subsection{Channel Pruning Based on the Dissimilarity of Angle}

Channel is a group of multidimensional data in CNN's convolutional kernels. AS is used to impose constrains on the model according to the difference. AS is widely used in natural language processing and data mining to measure cohesion within data clustering. AS $S\left ( A,B\right )$ of vectors $A$ and $B$ is demonstrated as follows:

\begin{equation}
    \mathit{S}(\mathit{A}, \mathit{B})=1-\left(\frac{\cos ^{-1}\left(f_{\text {similarity }}(A, B)\right)}{\pi}\right)
    \label{equ2}
\end{equation}
\begin{equation}
\begin{split}
    f_{\text {similarity}}(A, B)&=\cos (\theta)=\frac{A \cdot B}{\|A\|\|B\|}\\&=\frac{\sum_{i=1}^{n} A_{i} \times B_{i}}{\sqrt{\sum_{i=1}^{n}\left(A_{i}\right)^{2}} \times \sqrt{\sum_{i=1}^{n}\left(B_{i}\right)^{2}}}
    \label{equ3}
\end{split}
\end{equation}
where $A$ and $B$ are vectors to calculate AS $\mathit{A} _{i  }$ and $\mathit{B} _{i  }$ denote the component of vector $A$ and $B$ respectively.

We calculate the norm value of each convolutional kernel in the channel, and transform the channels and filters into the vectors. Then calculate the value of DACP according to AS:

\begin{equation}
    R_{c}=\sum_{l=1}^{\mathcal{L} } \sum_{i=1}^{c^{l}-1} \sum_{j=\mathrm{i}+1}^{c^{l}} \mathrm{~S}\left(X_{i}^{l}, X_{j}^{l}\right)
    \label{equ4}
\end{equation}
where $X_{i}^{l}$ and $X_{j}^{l}$  denotes the $\mathit{i}$-th and $\mathit{j}$-th channel vector of the $\mathit{l}$-th layer respectively.

\subsection{Loss function}\label{subsec3.4}

In supervised learning, $y$ represents the target tag. $C(x,W)$ represents the forward propagation result of input data $x$ in CNN. And $\mathcal{L}(\cdot)$ represents the target tag and the loss of output result. We use cross-entropy to calculate the fitting-error, and add DACP penalty $\mathcal{R} _{c}$ and LASSO penalty $\mathcal{R} _{g}$ to construct our channel pruning based on AD, which could be formulated as:

\begin{equation}
    \tilde{\mathcal{L}}(y, C(x, W))=\mathcal{L}(y, C(x, W))+\lambda \mathcal{R}_{c}+\beta \mathcal{R}_{g}
    \label{equ5}
\end{equation}
where $\lambda $ and $\beta$ denote the hyperparameter of the penalty term based on AD and LASSO penalty term respectively, which is conducive to adjust the level of penalty.

As for DACP penalty term, we use cosine-similarity instead of angle's. The larger the angle is, the smaller the similarity will be. The cosine-similarity among channels and filters plays a decisive role in our parameter penalty. The derivative of $f _{similarity{}}$ by the single vector $X _{i}$ is as follows:
\begin{equation}
\begin{split}
    \frac{\partial(f)}{\partial X_{i}}&=\frac{X_{j}\left(\left\|X_{i}\right\|\left\|X_{j}\right\|\right)-\left(X_{i}^{T} X_{j}\right) \frac{X_{i}\left\|X_{j}\right\|}{\left\|X_{i}\right\|}}{\left\|X_{i}\right\|^{2}\left\|X_{j}\right\|^{2}}\\&=\left(\frac{1}{\left(X_{i}^{T}\right)}-\frac{X_{i}}{\left\|X_{i}\right\|^{2}}\right) f_{\text {similarity }}
    \label{equ6}
\end{split}
\end{equation}
where the $f_{similarity}$ denotes the cosine-similarity of channel vector $X_{i}$ and $X_{j}$. Thus the gradient of penalty term with AD can be simplified as:
\begin{equation}
    \nabla_{w} \lambda \mathcal{R}_{\mathrm{c}} \propto\left(\left(\mathrm{W}^{T}\right)^{-1}-\frac{\mathrm{W}}{\|\mathrm{W}\|^{2}}\right) f_{\mathrm{w}}
    \label{equ7}
\end{equation}
where $f_{w} $ is the matrix of AS between vectors of channel.

It can be concluded from the gradient that the similarity between channels changes during the optimization processing. Combined with the constraints of GL, the similarity of multiple channels and filters with high similarity is penalized by AD. The similarity decreases gradually, and the norm value of some channels decreases sharply. After pruning, the model will become more compact, and generalization even improved.

\subsection{Simplify the Calculation of Similarity}

On VGG-16 model, the number of channels and filters reaches 512 after the 8-th layer, which means that the calculation of AS is huge. After calculating, the time complexity is $O\left(n^{3}\right)$. The larger the value of $n$ is, the larger the floating-point calculation resource is.

Considering the complexity to calculate AS among channels, we attempt to calculate the mean vector $B$ of channels and regard it as the base vector instead. We calculate AS between each channel and the base vector for approximation. Then the penalty term $\mathcal{R} _{c} $ of AD can be replaced by ${\mathcal{R} _{c} }' $ in Eq.\ref{equ8}.
\begin{equation}
    \mathcal{R}_{c}^{\prime}=\sum_{l=1}^{L} \sum_{i=1}^{c^{l}} \mathrm{~S}\left(X_{i}^{l}, B^{l}\right)
    \label{equ8}
\end{equation}
where $B^{l} $ denotes the mean vector of channel of the $\mathit{l}$-th layer. Meanwhile, with the calculating of the channel vector, the time complexity after approximation processing is $O\left(n^{2}\right)$, which seems not to be significantly improved. However, when calculating AS with the base vector. We can directly calculate AS in the form of tensor, and the actual calculation reduces.

\section{Experiments}

\subsection{Databases and Experimental Settings}

We evaluated our approach on VGG-16, and ResNet Deep Neural Networks (DNN) with the classical datasets: MNISIT, Caltech-101, CIFAR-10, and CIFAR-100.

We selected the following pruning methods that have worked well in both industry and academia to compare them with our experiments in turn. (1)Network Slimming (NS) \cite{bib19} applies L1 regularization to the scaling factor of batch normalization (BN) layer. L1 regularization inclines the scaling factor of BN layer to zero, so as to distinguish unimportant channels or neurons. (2) Soft Filter Pruning (SFP) \cite{bib9}, a dynamic pruning method, which can enable the pruning filter to participate in certain training to improve the efficiency of pruning. (3) Filter Pruning via Geometric Median (FPGM) \cite{bib4}, a network-compression method based on geometric center pruning. (4) Stripe-wise Pruning (SWP) \cite{bib20}. A learnable matrix is introduced to reflect the shape of each filter, and the matrix is used to guide model pruning.

Before model training, in addition to MNIST, we make data augmentation of random trimming and random mirror for other datasets. During the training, we have an arrangement for cosine learning rate decay. Finally, we use the decline rate of FLOPs as the pruning rate to evaluate the performance after pruning.

\subsection{Experiments on VGG-16}

\begin{table}[htbp]
    \caption{Performance of VGG-16.}
    \begin{center}
    \begin{threeparttable}
        \begin{tabular}{c|c c c}
            \toprule
            \textbf{Datasets} & \textbf{Method}\tnote{1} & \textbf{\makecell[c]{Pruned \\FLOPs(\%)}} & \textbf{\makecell[c]{Pruned \\Accuracy(\%)}} \\ \hline 
            \multirow{6}{*}{CIFAR-100} & baseline with L2\tnote{*} & 0 & \textbf{74.23} \\
                & L1\tnote{*} & 33.73 & 71.59 \\
                & NS \cite{bib19} & 37.1 & 72.09 \\
                & GL\tnote{*} & 39.84 & 72.18 \\
                & SFP \cite{bib9} & \textbf{41.8} & 70.28 \\
                & Ours & 42.18 & 73.12 \\ \hline
            \multirow{6}{*}{CIFAR-10} & baseline with L2\tnote{*} & 0 & \textbf{93.73} \\
                & L1\tnote{*} & 60.20 & 92.45 \\
                & NS \cite{bib19} & 51.0 & 92.8 \\
                & GL\tnote{*} & 66.57 & 92.51 \\
                & SWP \cite{bib20} & \textbf{71.16} & 92.85 \\
                & Ours & 66.86 & 93.31 \\ \hline
            \multirow{4}{*}{Caltech-101} & baseline with L2\tnote{*} & 0 & 95.28 \\
                & L1\tnote{*} & 20.3 & 95.1 \\
                & GL\tnote{*} & 16.3 & 93.5 \\
                & Ours & \textbf{22.95} & \textbf{95.55} \\
            \bottomrule
        \end{tabular}
        \begin{tablenotes}
            \item[1] In the ``Method'' column, ``baseline with L2'' means that L2 regularization is used in the model; ``Pruned FLOPs'' means that the model reduces the number of FLOPs to be lost.
            \item[*] The experiment is implemented by us.
        \end{tablenotes}
    \end{threeparttable}
    \end{center}
    \label{table:1}
\end{table}

VGG-16 is a 16-layer-single-branch CNN with 13 convolutional layers. We tested the performance of VGG-16 network on CIFAR-10, CIFAR-100 and Caltech-101 by using various pruning methods. Table \ref{table:1} reveals the relevant experiment results. Among them, NS \cite{bib19} and SFP \cite{bib9} are the data in the initial papers by others, and others are from our experiments.

\begin{table}[htbp]
    \caption{The influence of different level of penalty of GL on the dissimilarity of angle.}
    \begin{center}
    \begin{threeparttable}
        \begin{tabular}{c|c c}
        \toprule
        \textbf{Method} & \textbf{Pruned FLOPs (\%)} & \textbf{Pruned accuracy (\%)} \\ \hline
        ORI & 0 & $91.20 (\pm0.30)$ \\\
        L1 & $60.20 (\pm6.80)$ & $92.45 (\pm0.35)$ \\
        L2 & $0.06 (\pm0.06)$ & \textbf{93.73 ($\pm0.42$)} \\
        GL(1)\tnote{1} & $72.10 (\pm4.50)$ & $92.40 (\pm0.30)$ \\
        GL(2)\tnote{2} & $56.83 (\pm3.10)$ & $93.10 (\pm0.32)$ \\
        GL(1)\tnote{1} + ad\tnote{3} & \textbf{75.30 ($\pm$4.25)} & $92.81 (\pm0.44)$ \\
        GL(2)\tnote{2} + ad\tnote{3} & $66.86 (\pm4.50)$ & $93.31 (\pm0.26)$ \\ \hline
        \end{tabular}
        \begin{tablenotes}
            \item[1] Group LASSO with strong penalty.
            \item[2] Group LASSO with weak penalty.
            \item[3] Our AD penalty.
        \end{tablenotes}
    \end{threeparttable}
    \end{center}
    \label{table:2}
\end{table}

In order to analyze the influence of GL penalty-intensity on AD, we have done a lot of experiments on CIFAR-10. As shown in Table \ref{table:2}, the performance of AD on strong GL penalty is relatively stable in pruning and it generalizes better. When AD acts on the GL with weak penalty, its performance is stable in generalization and pruning improves.

\subsection{Experiments on ResNet}

ResNet is a multi-branch neural network structure composed of multiple residual blocks. Different from single-branch networks, pruning tends to cause a mismatch between the number of shortcuts filters for the residual block and the number of output filters. We solve this problem by taking their union.

From Table \ref{table:3}, we can compare the pruning of various methods on CIFAR-10 and CIFAR-100s. ResNet itself is a compact network. There is no obvious difference in accuracy between ours and others, but our method has distinct advantages in pruning rate. For shortcuts in ResNet reduce the number of effective filters in the residual blocks, and our approach screens out important filters from a similarity term.

\begin{table}[htbp]
    \caption{Performance of ResNet.}
    \begin{center}
    \begin{threeparttable}
        \begin{tabular}{c|c c c c}
        \toprule
            \textbf{Datasets} & \textbf{Model} & \textbf{Method} & \textbf{\makecell[c]{Pruned \\FLOPs(\%)}} & \textbf{\makecell[c]{Pruned \\Accuracy(\%)}} \\ \hline
            \multirow{16}{*}{\makecell[c]{CIFAR\\-10}} & \multirow{4}{*}{\makecell[c]{ResNet\\-18}} & baseline with L2\tnote{*} & 0.2 & \textbf{92.37} \\
                 &  & L1\tnote{*} & 22.62 & 90.55 \\
                 &  & GL\tnote{*} & 37.08 & 90.58 \\
                 &  & Ours & \textbf{52.5} & 90.51 \\ \cline{2-5}
                 & \multirow{6}{*}{\makecell[c]{ResNet\\-20}} & baseline with L2\tnote{*} & 0 & \textbf{92.63} \\
                 &  & L1\tnote{*} & 24.5 & 90.75 \\
                 &  & GL\tnote{*} & 40.26 & 90.56 \\
                 &  & SFP \cite{bib9} & 42.2 & 90.83 \\
                 &  & FPGM \cite{bib4} & 54 & 90.44 \\
                 &  & Ours & \textbf{62.5} & 90.61 \\ \cline{2-5}
                 & \multirow{6}{*}{\makecell[c]{ResNet\\-32}} & baseline with L2\tnote{*} & 0 & \textbf{93.65} \\
                 &  & L1\tnote{*} & 38.85 & 91.25 \\
                 &  & GL\tnote{*} & 42.65 & 91.64 \\
                 &  & SFP \cite{bib9} & 41.5 & 92.08 \\
                 &  & FPGM \cite{bib4} & 53.2 & 91.93 \\
                 &  & Ours & \textbf{58.46} & 91.76 \\ \cline{1-5}
            \multirow{8}{*}{\makecell[c]{CIFAR\\-100}} & \multirow{4}{*}{\makecell[c]{ResNet\\-18}} & baseline with L2\tnote{*} & 0 & \textbf{74.65} \\
                 &  & L1\tnote{*} & 21.6 & 73.02 \\
                 &  & GL\tnote{*} & 34.83 & 73.28 \\
                 &  & Ours & \textbf{43.16} & 73.24 \\ \cline{2-5}
                 & \multirow{4}{*}{\makecell[c]{ResNet\\-34}} & baseline with L2\tnote{*} & 0 & \textbf{74.86} \\
                 &  & L1\tnote{*} & 21.63 & 73.1 \\
                 &  & GL\tnote{*} & 48.11 & 72.47 \\
                 &  & Ours & \textbf{65.49} & 72.51 \\ \hline
        \end{tabular}
        \begin{tablenotes}
            \item[*] The experiment is implemented by us. Some methods do not show the best performance due to the influence of hyperparameters, but we try our might to get the best results and then conduct correlative experiments.
        \end{tablenotes}
    \end{threeparttable}
    \end{center}
    \label{table:3}
\end{table}

\subsection{Feature Visualization}

\begin{figure}[htbp]
    \subfigure[Input]{
        \begin{minipage}[t]{.28\linewidth}
            \centering
            \includegraphics[scale=0.22]{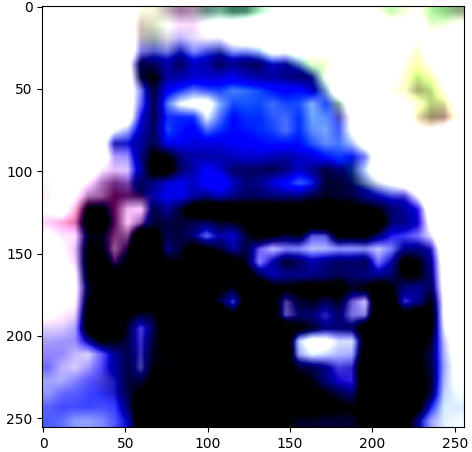}
            \label{3a}
        \end{minipage}
    }
    \subfigure[GL]{
         \begin{minipage}[t]{.28\linewidth}
            \centering
            \includegraphics[scale=0.1]{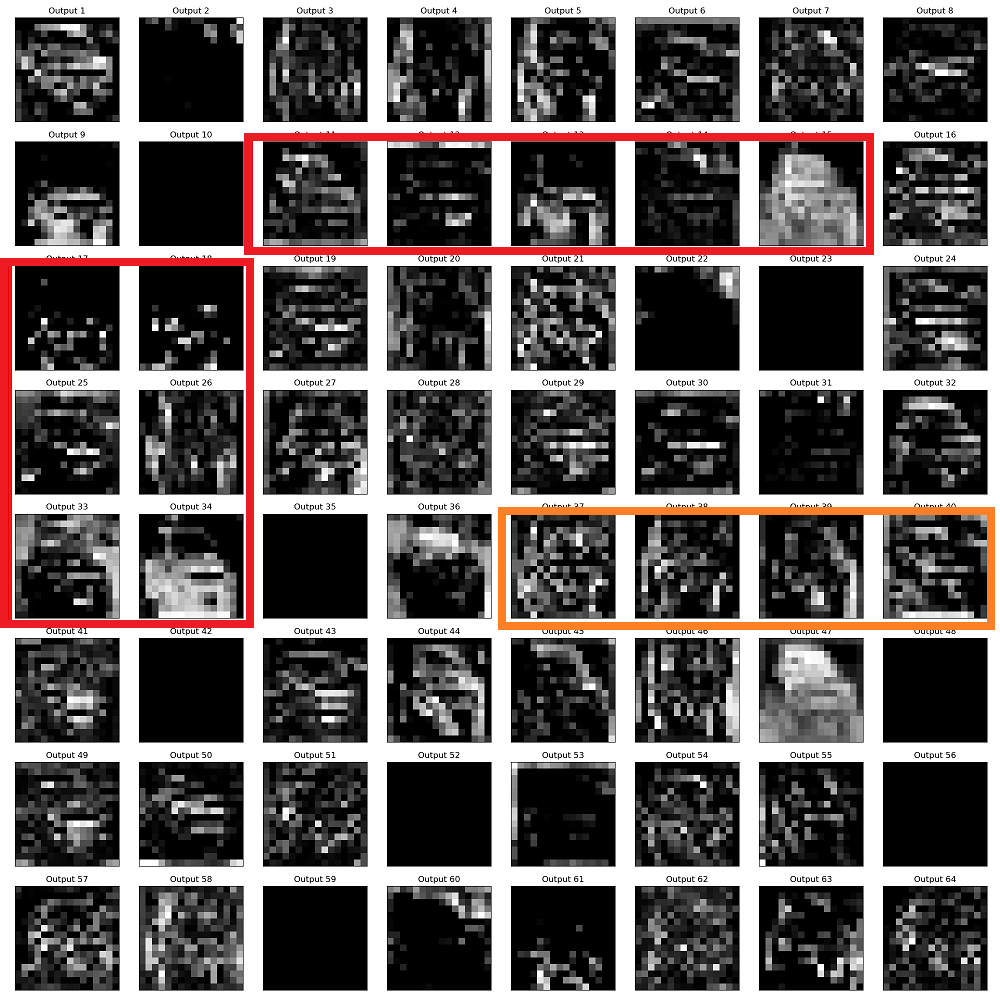}
            \label{3b}
        \end{minipage}
    }
    \subfigure[DACP]{
        \begin{minipage}[t]{.28\linewidth}
            \centering
            \includegraphics[scale=0.1]{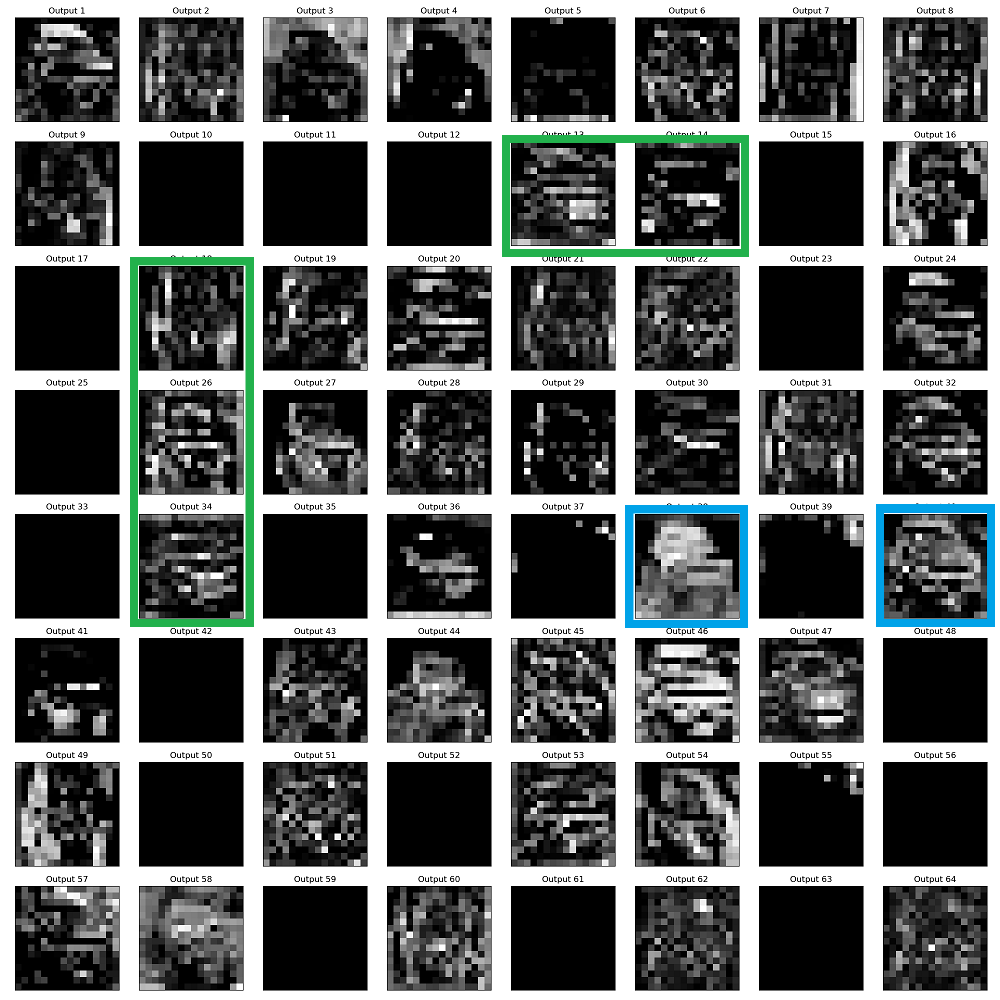}
            \label{3c}
        \end{minipage}
    }
\caption{Feature visualization. There are 64 filters in the second layer of VGG-16. Totally black blocks in the figure indicate that the filters have been cut off.}
\label{figure3}
\end{figure}

Further, we realize visualization analysis of output features in the model. On CIFAR-10, make visualization analysis and comparation on the output feature of second convolutional kernel on VGG-16, which is shown in Fig. \ref{figure3}. We select a car picture as the input feature. After the convolutional operation of convolutional kernel of the second layer, we intercepted the output feature of GL method and DACP, as depicted in Fig. \ref{figure3}\subref{3b} and Fig. \ref{figure3}\subref{3c} respectively. It is obvious that GL has a higher similarity of adjacent features, while AD has a more well-stacked feature map. The feature map of the red box in Fig. \ref{figure3}\subref{3b} can be replaced by the feature map in green box in Fig. \ref{figure3}\subref{3c} after AD penalty is applied. The features of the orange box in Fig. \ref{figure3}\subref{3b} are changed to those of the blue box in Fig. \ref{figure3}\subref{3c}, which can be regarded as the restoration of filters.

\section{Conclusion}

In the experiment, it can be seen that 3D-filters show different levels of clustering while applying different level of LASSO penalties. Compared with L1 penalty, GL penalty has better clustering effect. The distribution of 3D-filters is relatively uniform, and the degree of model simplification is higher. Combined with this circumstance, DACP is proposed. The relationship of channels and filters in convolutional kernel is discussed, and the corresponding DACP is constructed with cosine similarity. Experiment results show that the DACP can make the 3D-filter clustering in the model more uniform and improve the sparsity of the model.

In this paper, the calculation of DACP adopts basis vector approximation, which will be more convictive if a more suitable way can be found to express it. In the future, we will start with the connectivity-power that mentioned in the beginning of this article. We attempt to guide model sparsity based on connectivity-power, and play a role of regularization to push the performance to a higher stage.


\end{document}